\documentclass{article}
\usepackage{spconf}
\usepackage{cite}
\usepackage{xurl}
\usepackage{hyperref}
\usepackage{graphicx}
\usepackage{textcomp}
\usepackage{xcolor}
\usepackage{url}
\usepackage{booktabs}
\usepackage{amsfonts}
\usepackage{nicefrac}
\usepackage{microtype}
\usepackage{subcaption}
\usepackage{array}
\usepackage{algorithm}
\usepackage{algorithmicx}
\usepackage{algpseudocode}
\usepackage{amsmath}
\usepackage{amssymb}
\usepackage{mathtools}
\usepackage{amsthm}
\usepackage{extpfeil}
\usepackage{multirow}
\usepackage{tikz}
\usepackage{makecell}
\usepackage{siunitx}
\usepackage{thmtools}
\usepackage{mathrsfs}
\usepackage{tabularx}
\usepackage[most]{tcolorbox}
\usepackage{enumerate}
\usepackage{paralist}
\usepackage{enumitem}

\usetikzlibrary{quantikz2}
\usetikzlibrary{calc, decorations.pathreplacing}
\usetikzlibrary{decorations.pathreplacing}
\usetikzlibrary{positioning, fit, backgrounds, arrows.meta}
\colorlet{myborder}{blue!55!black}   
\colorlet{myblock}{blue!20}          
\colorlet{mycard}{green!20}          
\colorlet{mytitle}{black}            
\colorlet{myedge}{blue!55!black}     
\definecolor{Medium Blue}{rgb}{0, 0, 0.8}

\DeclareMathOperator{\A}{\mathcal{A}}
\DeclareMathOperator{\I}{\mathcal{I}}
\DeclareMathOperator{\R}{\mathbb{R}}

\DeclarePairedDelimiterX{\norm}[1]{\lVert}{\rVert}{#1}

\theoremstyle{definition}

\theoremstyle{remark}

\title{Quantum Hierarchical Reinforcement Learning via Variational Quantum Circuits}
\name{
  Yu-Ting Lee$^{1}$ \quad
  Samuel Yen-Chi Chen$^{2}$\thanks{The views expressed in this article are those of the authors and do not represent the views of Wells Fargo. This article is for informational purposes only. Nothing contained in this article should be construed as investment advice. Wells Fargo makes no express or implied warranties and expressly disclaims all legal, tax, and accounting implications related to this article.} 
}
\address{
  $^{1}$Graduate Institute of Communication Engineering, National Taiwan University, Taipei, Taiwan \\
  $^{2}$Wells Fargo, New York, NY, USA \\
  r14942088@ntu.edu.tw, yen-chi.chen@wellsfargo.com
  }
\begin{document}
\maketitle
\begin{abstract}
While parameterized quantum computations have shown success in standard reinforcement learning (RL), whether these advantages adapt to hierarchical RL (HRL) remains a critical open question. This work demonstrates that variational quantum circuits (VQCs) can effectively enhance HRL agents based on the option-critic architecture. Evaluated in standard environments, a hybrid HRL agent with a quantum feature extractor outperforms classical baselines while using fewer parameters. We also identify an architectural bottleneck: using VQCs for option-value estimation severely degrades learning. Further ablations reveal how quantum circuit design affects performance. Our work establishes design principles for parameter-efficient hybrid HRL agents.
\end{abstract}

\section{Introduction}
Reinforcement learning (RL) has achieved great success in solving decision-making problems~\cite{Silver2017AlphaGoZero, Schrittwieser2020, pmlr-v48-mniha16}. However, standard RL algorithms often struggle with long-horizon tasks and sparse rewards~\cite{NIPS2017_453fadbd}. Hierarchical reinforcement learning (HRL) mitigates these issues through temporal abstraction, representing knowledge as courses of action spanning multiple time scales~\cite{NIPS1997_5ca3e9b1, 10.5555/1622262.1622268, SUTTON1999181, Bacon_Harb_Precup_2017, 10.1007/3-540-45622-8_16}. A prominent example is the \emph{option-critic architecture}~\cite{Bacon_Harb_Precup_2017}, which enables an end-to-end, gradient-based learning of temporal abstraction called \emph{options}.

Current HRL research remains predominantly classical. In standard non-hierarchical settings, quantum reinforcement learning (QRL) has shown promise in achieving performance comparable to or exceeding classical approaches with significantly fewer parameters~\cite{Skolik2022quantumagentsingym, 10.1109/TSMCB.2008.925743,meyer2024surveyquantumreinforcementlearning, Lockwood_Si_2020, NEURIPS2021_eec96a7f, icaart24_kolle, Chen:2020opi, Kwak:2021quh, CHEN2023321, Caro2022}. Whether HRL can be similarly enhanced by leveraging the expressivity inherent in quantum computing~\cite{NEURIPS2021_eec96a7f, Expressibility_and_Entangling, PerezSalinas2020datareuploading, PhysRevA.103.032430} remains a critical and largely unexplored open question.

In this work, we introduce a hybrid HRL agent by integrating variational quantum circuits (VQCs) into the option-critic framework.\footnote{Code is available at \url{https://github.com/Yu-TingLee/quantum-option-critic}.} We substitute classical components\textemdash feature extractors, option-value functions, termination functions, and intra-option policies\textemdash with VQCs. Our modular approach lets us evaluate the distinct impacts of quantum components on the learning dynamics of hybrid HRL agents. Our contributions:
\begin{itemize}
    \item[(1)] We show that a hybrid HRL agent with a quantum feature extractor is parameter-efficient and outperforms classical baselines.
    \item[(2)] We identify a critical bottleneck: using VQCs for option-value function approximation fails to provide useful learning signals, causing learning failure.
    \item[(3)] Ablation studies show how architectural choices, specifically circuit depth, trainable input scaling, and entanglement, affect the hybrid agent's efficacy.
\end{itemize}

\vspace{-1em}
\section{Preliminaries}\label{sec:preliminaries}
\subsection{The Option-Critic Architecture}

We operate within a standard Markov decision process (MDP) extended by the options framework. An MDP is defined by a state space $\mathcal{S}$, an action space $\A$, a transition probability $P\colon \mathcal{S} \times \A \to \Delta(\mathcal{S})$, a reward function $r \colon \mathcal{S} \times \A \to \R$, and a discount factor $\gamma \in [0,1)$. A Markovian option $\omega \in \Omega$ is a triple $(\I_\omega, \pi_\omega, \beta_\omega)$ in which $\I_\omega \subset \mathcal{S}$ is an initiation set, $\pi_\omega$ is an intra-option policy, and $\beta_\omega \colon \mathcal{S} \to [0,1]$ is a termination function. Under a call-and-return workflow, an agent selects an option $\omega$, executes actions according to $\pi_\omega$ until termination is triggered by $\beta_\omega$, and then repeats the process. The option-critic architecture~\cite{Bacon_Harb_Precup_2017} enables an end-to-end learning and discovery of options using policy gradient results, where an option-value function $Q_{\Omega}(s,\omega)$ and a state-value function $V_{\Omega}(s)$ are used to evaluate the gradient updates.

\vspace{-0.5em}
\subsection{Variational Quantum Circuits}
Variational quantum circuits (VQCs) are trainable quantum models for processing classical data. A standard VQC consists of three components. First, an encoding circuit $U(s)$ encodes a classical state $s$ into an $n$-qubit system, yielding an encoded state $U(s)|0\rangle^{\otimes n}$, where $|0\rangle^{\otimes n}$ is the ground state of the quantum system. Then, a parameterized circuit $V(\theta)$ evolves this state into $V(\theta)U(s)|0\rangle^{\otimes n}$. Here, $V(\theta)$ typically consists of alternating layers of trainable single-qubit rotations and multi-qubit entangling gates. Finally, classical information is extracted by evaluating the expectation values of predefined Hermitian observables $\{O_k\}_{k=1}^K$, such as Pauli-Z matrices. The computation of a VQC can be summarized as a quantum function $f(s;\theta) = (\langle O_1\rangle, \dots, \langle O_K\rangle)$ where $
\langle O_k\rangle =
\langle 0|^{\otimes n} U^\dagger(s) V^\dagger(\theta) O_k V(\theta) U(s) |0\rangle^{\otimes n}.$

\vspace{-0.5em}
\section{Option-Critic Algorithm}\label{sec:option_critic_algorithm}

Since the option-critic architecture~\cite{Bacon_Harb_Precup_2017} is agnostic to any specific learning algorithm, we briefly outline our implementation. The agent maintains a main network ($\theta$) and a target network ($\theta'$), where each network parameterizes an option-value function $Q_\Omega$ and $|\Omega|$ options comprising intra-option policies $\pi_\omega$ and termination functions $\beta_\omega$. Options are selected via an $\epsilon$-greedy policy over the option-values $Q_\Omega(s,\cdot\mid \theta)$ of the current state.
At each step, the agent executes an action from $\pi_\omega$ and stores the resulting transition  $(s,\omega,r,s')$ in a replay buffer.

Learning relies on a unified loss with online actor updates and periodic off-policy critic updates. The actor is optimized using a one-step TD target $y_\mathrm{actor}$, computed using the slow target network. This drives two components: (i) a policy loss using $(y_\mathrm{actor} - Q_\Omega(s,\omega\mid\theta))$ as an advantage, and (ii) a termination loss encouraging option switching if the active option's value falls below $\max_{\omega'} Q_\Omega(s,\omega'\mid\theta)$.  The critic is updated periodically by sampling mini-batches from the buffer to compute an MSE loss against batch TD targets. This MSE is directly added to total loss, allowing the optimizer to update the parameters $\theta$ end-to-end. Finally, option termination is triggered by $\beta_\omega(s'\mid \theta)$, and $\theta'$ is synchronized periodically.

\vspace{-0.5em}
\section{VQCs for Hybrid Option-Critic}\label{sec:vqc_hybrid_option_critic}
\begin{figure}[t!]
\centering
\resizebox{\columnwidth}{!}{%
\begin{tikzpicture}[
    >=Stealth, font=\sffamily\scriptsize,
    blk/.style={rectangle, draw=black!80, thick, fill=blue!20, fill opacity=0.45,
                text opacity=1, text width=2cm, align=center, rounded corners,
                minimum height=0.65cm, text=black},
    crd/.style={blk, fill=green!20},
    grp/.style={draw=black!80, dashed, thick, rounded corners, inner sep=5pt},
    ttl/.style={text=black, font=\sffamily\scriptsize, fill=white, inner sep=1.5pt},
    ar/.style={->, thick, draw=black!80},
    ln/.style={thick, draw=black!80}
]
\node[crd] (input) {Input obs.\\(B,in)};
\node[blk, below=0.8cm of input] (feat) {NN / VQC\\(B,in)$\to$(B,in)};

\node[blk, below=1.35cm of feat, xshift=-3.2cm] (qh) {NN / VQC\\(B,in)$\to$(B,$|\Omega|$)};
\node[crd, below=0.4cm of qh] (qo) {Logits (B,$|\Omega|$)};

\node[blk, below=1.35cm of feat] (th) {NN / VQC\\(B,in)$\to$(B,$|\Omega|$)};
\node[crd, below=0.4cm of th] (ts) {Sigmoid};
\node[crd, below=0.4cm of ts] (to) {Logits (B,$|\Omega|$)};

\node[blk, below=1.35cm of feat, xshift=3.2cm] (ph) {NN / VQC\\(B,in)$\to$(B,out)};
\node[crd, below=0.4cm of ph] (ps) {Softmax};
\node[crd, below=0.4cm of ps] (po) {Action probs.\\(B,out)};

\node[grp, fit=(feat)] (gf) {};
\node[grp, fit=(qh)(qo)] (gq) {};
\node[grp, fit=(th)(ts)(to)] (gt) {};
\node[grp, fit=(ph)(ps)(po)] (gp) {};

\coordinate (m) at ($(feat.south)+(0,-0.5)$);
\draw[ar] (input) -- (gf.north);
\draw[ln] (feat.south) -- (m);
\draw[ar] (m) -| (gq.north);
\draw[ar] (m) -- (gt.north);
\draw[ar] (m) -| (gp.north);
\draw[ar] (qh) -- (qo);
\draw[ar] (th) -- (ts);  \draw[ar] (ts) -- (to);
\draw[ar] (ph) -- (ps);  \draw[ar] (ps) -- (po);

\node[ttl, above=4pt of gf] {Feature Extractor};
\node[ttl, above=4pt of gq] {Option-Value};
\node[ttl, above=4pt of gt] {Termination};
\node[ttl, above=4pt of gp] {Intra-Option Policies ($\times|\Omega|$)};
\end{tikzpicture}}
\caption{\textbf{Model architecture for hybrid option-critic.}}
\label{fig:model_architecture}
\end{figure}

\begin{table}[!t]
  \centering
  \caption{\textbf{Model configurations.} Downstream parameters:
  Option-Value / Termination / Intra-Option.}
  \label{tab:architecture}
  \ninept
  \resizebox{\columnwidth}{!}{
  \setlength{\tabcolsep}{4pt}
  \begin{tabular}{@{}lll l r@{}}
    \toprule
    \textbf{Env.} & \textbf{Component} & \textbf{Type} & \textbf{Architecture} & \textbf{Parameters} \\
    \midrule
    \multirow{4}{*}{CartPole}
      & \multirow{2}{*}{Feat. Extract.} & Classical & MLP (8 neurons) & 76 \\
      &                              & Quantum   & VQC (4 layers)  & 48 \\
      \cmidrule(l){2-5}
      & \multirow{2}{*}{Downstream}  & Classical & Linear layers   & 10 / 10 / 20 \\
      &                              & Quantum   & VQC (1 layer)   & 12 / 12 / 24 \\
    \midrule
    \multirow{4}{*}{Acrobot}
      & \multirow{2}{*}{Feat. Extract.} & Classical & MLP (8 neurons) & 110 \\
      &                              & Quantum   & VQC (5 layers)  & 90 \\
      \cmidrule(l){2-5}
      & \multirow{2}{*}{Downstream}  & Classical & MLP (3 neurons) & 29 / 29 / 66 \\
      &                              & Quantum   & VQC (2 layers)  & 36 / 36 / 72 \\
    \bottomrule
  \end{tabular}
  }
\end{table}
Our hybrid option-critic agent comprises four components: a shared feature extractor, an option-value function, a termination function, and a set of intra-option policies. For parameter efficiency, the option-value function and termination functions each output values for all options simultaneously. See Fig.~\ref{fig:model_architecture} for an illustration of our model architecture.

We adopt a data re-uploading architecture for VQC~\cite{NEURIPS2021_eec96a7f, PhysRevA.103.032430}. Each of the $L$ layers consists of $R_x$ encoding gates with trainable input scaling $\lambda$, entangling CNOT gates, and parameterized $R_y$ and $R_z$ rotations. For measurement, we utilize Pauli-Z observables for each qubit. All inputs are normalized to the range $[-\pi, \pi]$. For feature extractor VQCs, unbounded classical observations $o_\mathrm{unbound}$ are mapped via $2 \arctan(o_\mathrm{unbound})$, while variables symmetrically bounded in $[-c, c]$ are scaled by $\frac{\pi}{c}$. For downstream VQCs, the input feature vector $h$ is uniformly transformed via $2 \arctan(h)$.
\begin{figure*}[!th]
    \centering
    \resizebox{1\linewidth}{!}{%
        \includegraphics{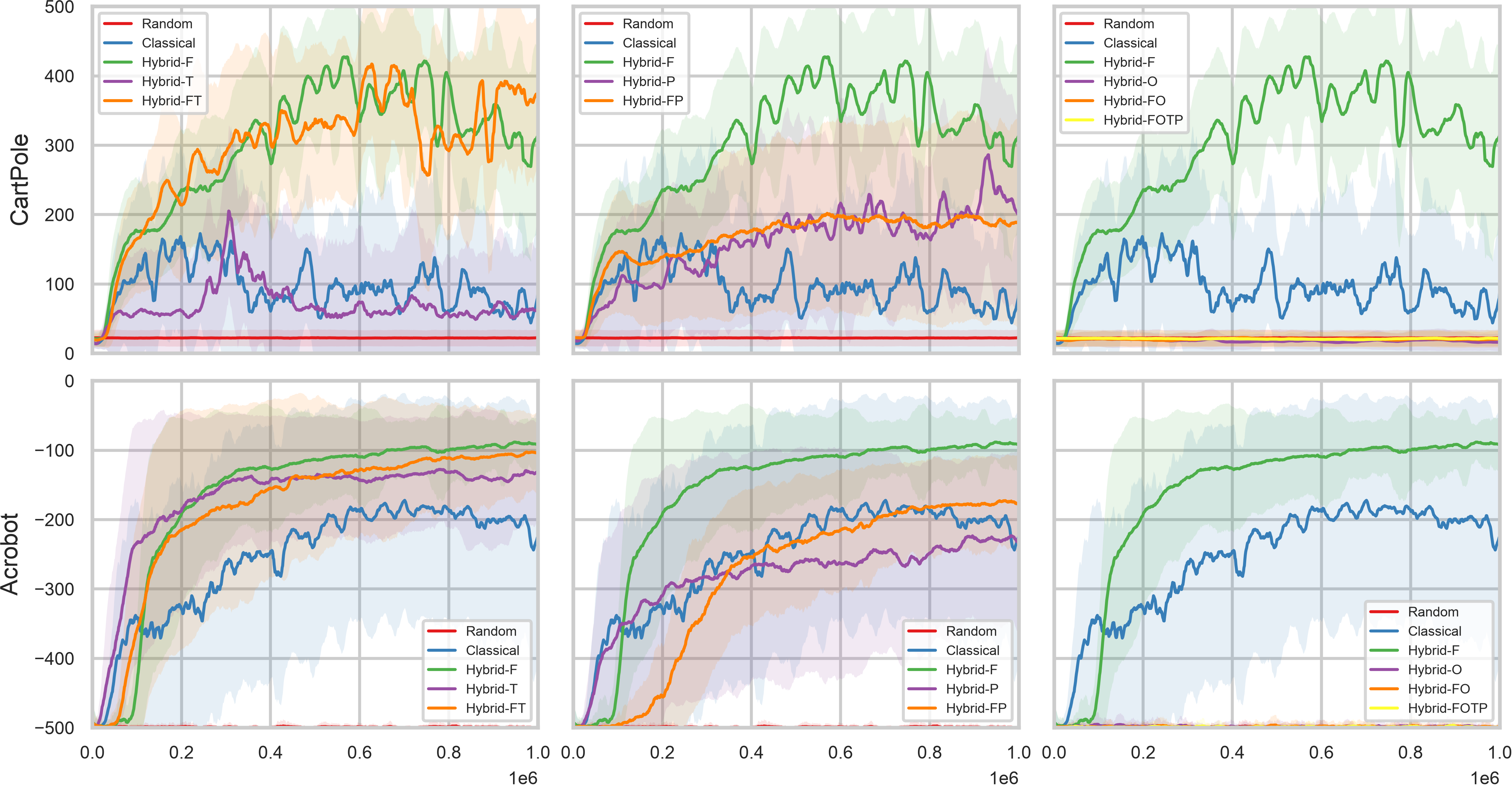}%
    }
    \caption{\textbf{Performance of hybrid option-critic.} Curves represent mean episodic reward vs. total steps. Shaded area is standard deviation. Each curve is averaged with a trailing window of 2000 steps and averaged over 10 runs with different random seeds.}\label{fig:main_result}
\end{figure*}

\vspace{-0.5em}
\section{Experimental Settings}\label{sec:exp_settings}

We define eight hybrid variants by substituting classical components with VQCs. Substituted components are labeled \textbf{F} (Feature extractor), \textbf{O} (Option-value function), \textbf{T} (Termination function), and \textbf{P} (intra-option Policies). Detailed configurations are in Table~\ref{tab:architecture}.

We consider two classical environments~\cite{towers2025gymnasiumstandardinterfacereinforcement}: CartPole and Acrobot, which have continuous state spaces and discrete action spaces.\footnote{\url{https://gymnasium.farama.org/}} Our choice of environments is consistent with prior QRL work~\cite{Lockwood_Si_2020, NEURIPS2021_eec96a7f,Chen:2020opi,Skolik2022quantumagentsingym}. The hybrid variants are compared against a classical baseline and a random baseline (uniform action sampling). All models are trained using Adam with a learning rate of 0.0005, a discount factor $\gamma=0.99$, and an exploration rate decaying exponentially from 1.0 to 0.05. Target networks synchronize every 200 steps.

\vspace{-1em}
\section{Experiments}\label{sec:exp}
Fig.~\ref{fig:main_result} presents the results of the eight hybrid models and the two baselines. Further analysis and ablation studies are presented in Fig.~\ref{fig:analysis_option_value}, Fig.~\ref{fig:more_options}, and Fig.~\ref{fig:architectural_choices}.

\vspace{-0.5em}
\subsection{Advantage of the Quantum Feature Extractor}
Hybrid-F and Hybrid-FT are the strongest across both environments, and among single substitutions, Hybrid-F is the most effective. This advantage likely stems from the expressivity of VQCs when processing classical observations. Hybrid-F is also parameter-efficient, cutting the parameter count by roughly 24\% in CartPole and 8.5\% in Acrobot. In isolation, quantum downstream components yield only task-specific gains: Hybrid-T improves on the classical baseline only in Acrobot, and Hybrid-P only in CartPole. Pairing the termination function with the feature extractor removes this limitation: Hybrid-FT matches Hybrid-F in both environments. The quantum feature extractor is thus not only strong on its own but also remains effective when paired with the termination function.

\subsection{Performance Degradation from the Quantum Option-Value Function}
\begin{figure}[!th]
    \centering
    \includegraphics[width=0.94\linewidth]{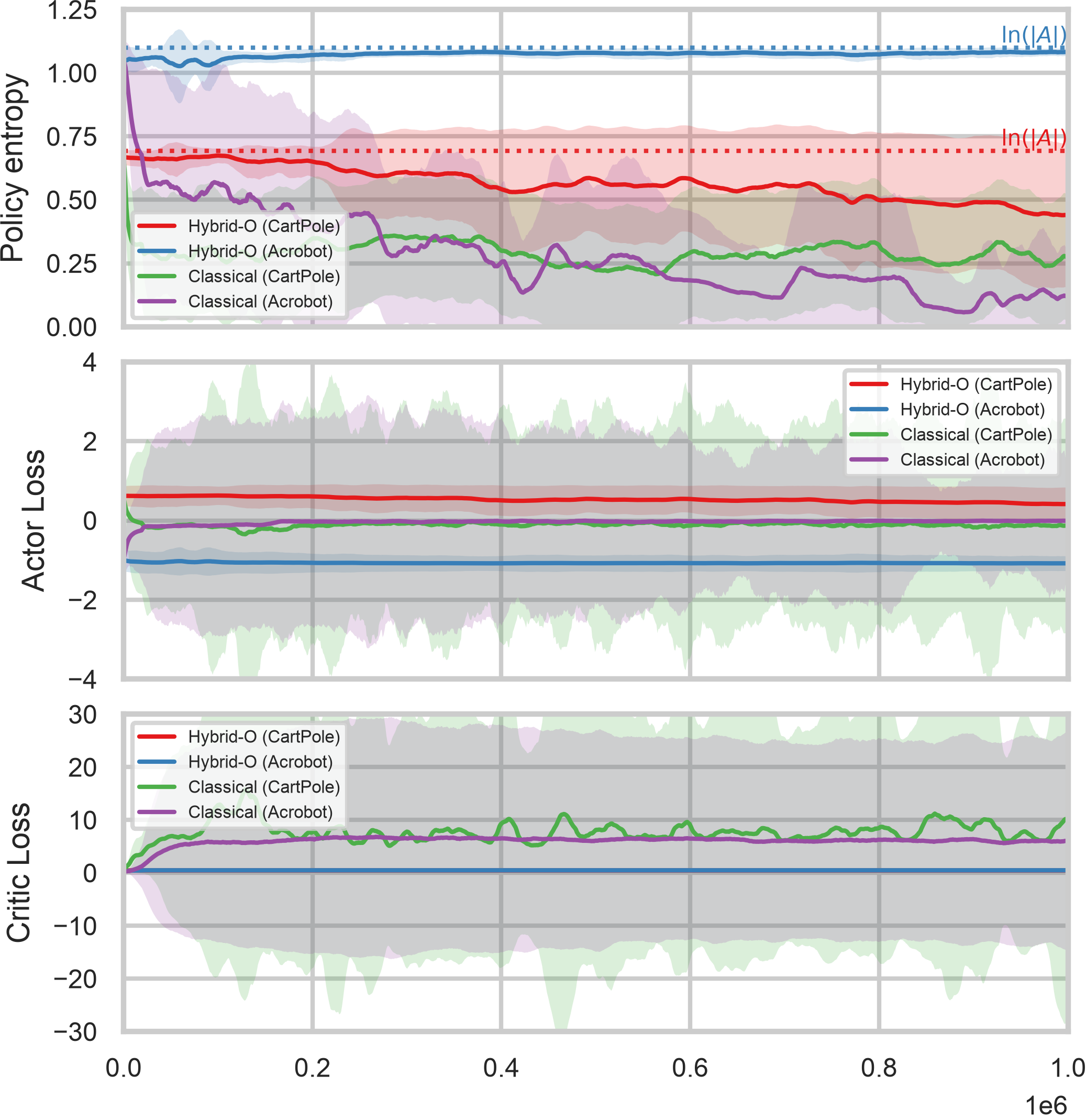}
    \caption{\textbf{Analysis of the option-value bottleneck.} We compare policy entropy, actor loss, and critic loss between Hybrid-O and the classical baselines in both environments.}\label{fig:analysis_option_value}
\end{figure}

Replacing the option-value function with a VQC severely degrades performance, failing to surpass random baselines in either environments; even with the feature extractor, Hybrid-FO and Hybrid-FOTP collapse alongside Hybrid-O. From Fig.~\ref{fig:analysis_option_value}, the learning dynamics reveal that Hybrid-O exhibits flat actor and critic losses alongside policy entropy near the theoretical maximum of $\ln(|\A|)$, effectively resulting in uniform action sampling. This suggests the quantum critic fails to provide useful learning signals. A promising direction is to investigate whether trainable observables~\cite{11249836} can resolve this bottleneck.

\subsection{Impact of the Number of Options}
\begin{figure}[!th]
    \centering
    \includegraphics[width=0.94\linewidth]{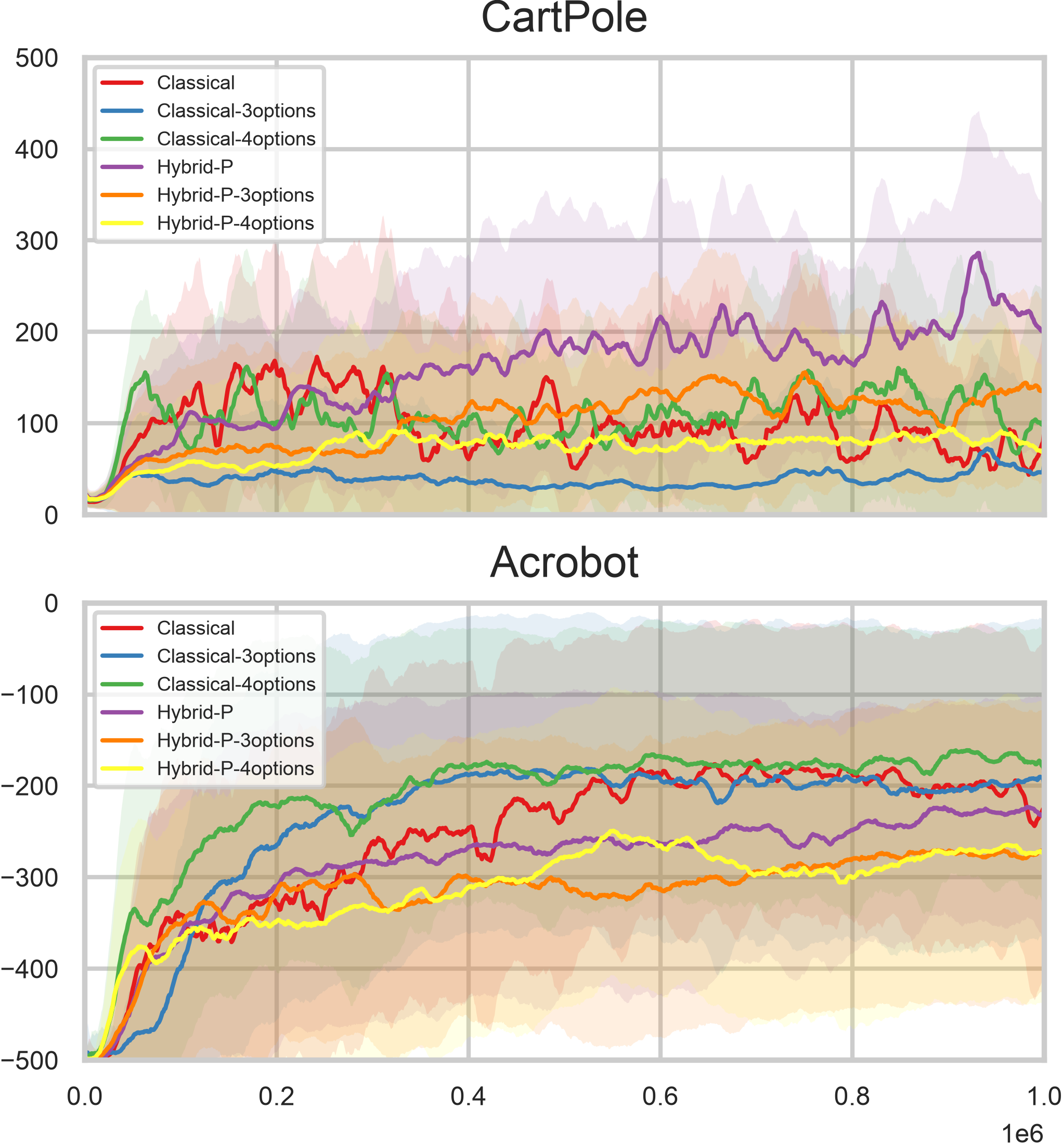}
    \caption{\textbf{Impact of the number of options on performance.} We report learning curves of classical and quantum intra-option policies with increased temporal abstraction.}\label{fig:more_options}
\end{figure}

Expanding the number of options beyond $|\Omega|=2$ to 3 or 4 does not yield a consistent pattern of improvement or degradation for classical or quantum policies (Fig.~\ref{fig:more_options}). This indicates that simply increasing available temporal abstraction does not reliably enhance agents in these specific environments. 

\subsection{Impact of Architectural Choices}
To evaluate how specific architectural choices influence performance, we conduct ablation studies on Hybrid-F. We vary one component at a time: incrementing or decrementing VQC depth by 2 layers, fixing input scaling parameters $\lambda=1$, or removing entangling CNOT gates (Fig.~\ref{fig:architectural_choices}). We can make the following observations:

\begin{figure}[!ht]
    \centering
    \includegraphics[width=0.92\linewidth]{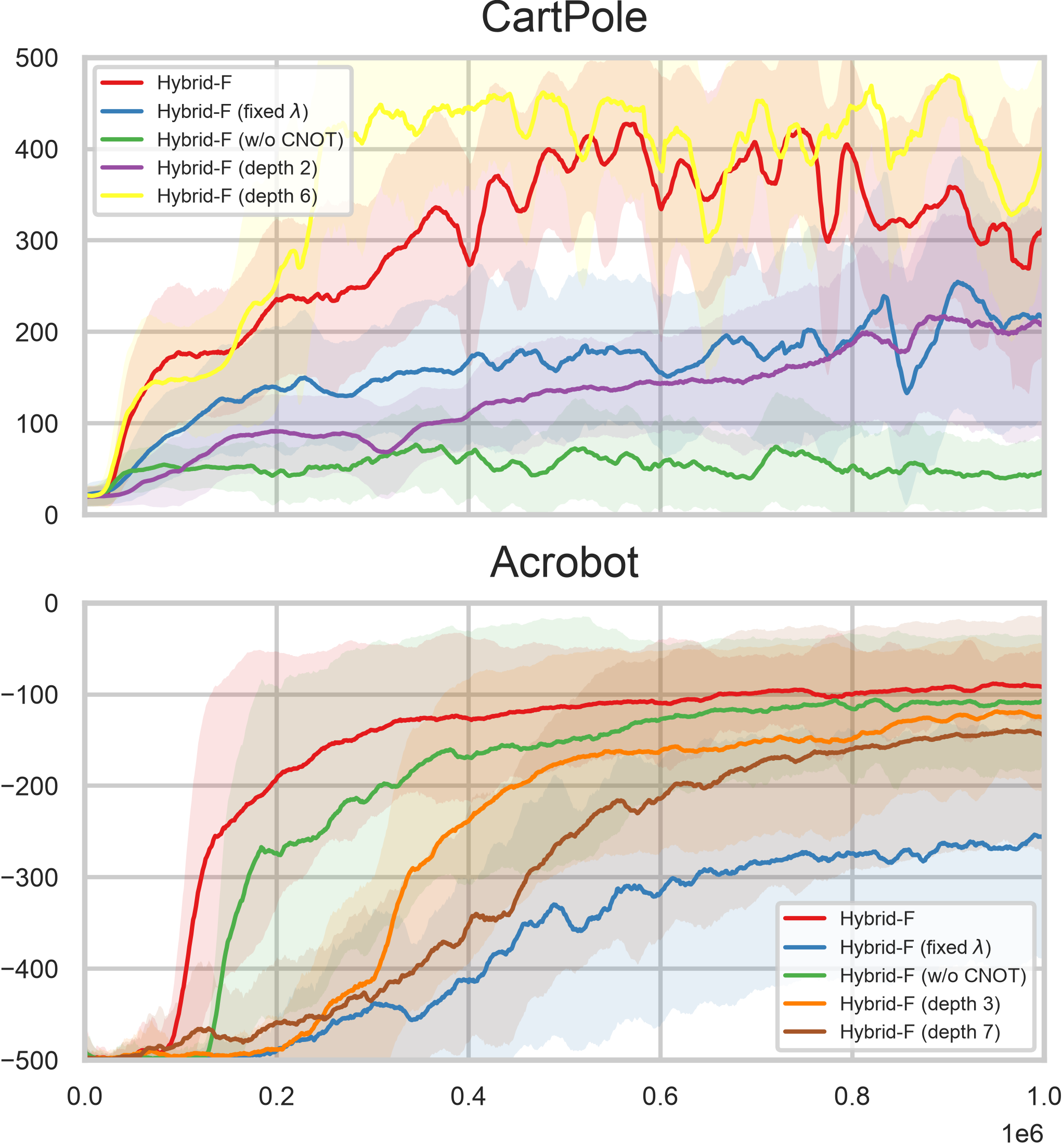}
    \caption{\textbf{Influence of Architectural Choices.} We ablate Hybrid-F by (i) increasing or decreasing VQC depth by 2 layers, (ii) fixing the input scaling $\lambda$ to 1, or (iii) removing CNOT gates.}\label{fig:architectural_choices}
\end{figure}

\begin{itemize}
    \item[(1)] \textbf{Influence of model depth:} Increasing VQC depth does not consistently improve performance, while decreasing depth degrades performance in both environments. 
    \item[(2)] \textbf{Influence of scaling parameters $\boldsymbol\lambda$:} Trainable input scaling parameters $\lambda$ greatly benefit the hybrid agents.
    \item[(3)] \textbf{Influence of entanglement:} Removing the CNOT gates degrades performance in both environments, confirming that entanglement is crucial for the hybrid agent.
\end{itemize}

Overall, our results show that VQC placement within the hierarchy is a decisive factor, and the quantum feature extractor is consistently strong and parameter-efficient; in contrast, the quantum option-value function fails to learn, leaving policies near-uniform. Ablations show that circuit depth, entanglement, and trainable input scaling remain critical.

\section{Conclusion}\label{sec:conclusion}
This work provides a proof-of-principle demonstration that VQCs can successfully enhance HRL. Our key findings are threefold: (1) a quantum feature extractor enables a hybrid HRL agent to outperform classical baselines with fewer parameters, (2) using a VQC for the option-value function causes severe learning failure, exposing a critical architectural bottleneck, and (3) circuit depth, entanglement, and trainable input scaling are important factors for hybrid agents. Together, these results provide architectural guidelines for integrating quantum circuits into HRL agents.


\vfill\pagebreak

\ninept
\bibliographystyle{IEEEbib}
\bibliography{references}

\end{document}